\documentclass{bmvc2k}
\usepackage{times}
\usepackage{epsfig}
\usepackage{graphicx}
\usepackage{amsmath}
\usepackage{amssymb}
\usepackage{amsthm}
\usepackage{algorithm}
\usepackage{algorithmic}
\usepackage{caption}
\usepackage{bm}
\usepackage{array}
\usepackage{color}
\usepackage{url}

\title{Jointly Learning Non-negative Projection and Dictionary with Discriminative Graph Constraints for Classification}

\addauthor{Weiyang Liu}{wyliu@pku.edu.cn}{12}
\addauthor{Zhiding Yu}{yzhiding@andrew.cmu.edu}{3}
\addauthor{Yandong Wen}{yandongw@andrew.cmu.edu}{3}
\addauthor{Rongmei Lin}{rongmei.lin@emory.edu}{4}
\addauthor{Meng Yang*}{yang.meng@szu.edu.cn}{1}

\addinstitution{
 College of Computer Science \& Software Engineering,\\
 Shenzhen University, China
}
\addinstitution{
 School of ECE,\\
 Peking University, China
}
\addinstitution{
 Dept. of ECE,\\
 Carnegie Mellon University, USA
}
\addinstitution{
 Dept. of Math \& Computer Science,\\
 Emory University, USA
}

\runninghead{W. Liu et al.}{Jointly Learning Non-negative Projection and Dictionary}

\begin{document}

\maketitle

\begin{abstract}
Sparse coding with dictionary learning (DL) has shown excellent classification performance. Despite the considerable number of existing works, how to obtain features on top of which dictionaries can be better learned remains an open and interesting question. Many current prevailing DL methods directly adopt well-performing crafted features. While such strategy may empirically work well, it ignores certain intrinsic relationship between dictionaries and features. We propose a framework where features and dictionaries are jointly learned and optimized. The framework, named ``joint non-negative projection and dictionary learning'' (JNPDL), enables interaction between the input features and the dictionaries. The non-negative projection leads to discriminative parts-based object features while DL seeks a more suitable representation. Discriminative graph constraints are further imposed to simultaneously maximize intra-class compactness and inter-class separability. Experiments on both image and image set classification show the excellent performance of JNPDL by outperforming several state-of-the-art approaches.
\end{abstract}

\section{Introduction}
Sparse coding has been widely applied in a variety of computer vision problems where one seeks to represent a signal as a sparse linear combination of bases (dictionary atoms). Dictionary plays an important role as it is expected to robustly represent components of the query signal. \cite{wright2009robust} proposed the sparse representation-based classification (SRC) in which the entire training set is treated as a structured dictionary. Methods taking off-the-shelf bases (e.g., wavelets) as the dictionary were also proposed \cite{huang2006sparse}. While such strategy is straight forward and convenient, research also indicates that it may not be optimal for classification tasks. Many currently prevailing approaches choose to learn dictionaries directly from training data and have leaded to good results in face recognition \cite{zhang2010discriminative,jiang2013label,kong2012dictionary,yang2011fisher,yang2014sparse} and image classification \cite{yang2014latent}. These dictionary updating strategies are referred to as dictionary learning (DL). DL received significant attention for its excellent representation power. Such advantage mainly comes from the fact that allowing the update of dictionary atoms often results in additional flexibility to discover better intrinsic signal patterns, therefore leading to more robust representations.

\begin{figure}[t]
  \renewcommand{\captionlabelfont}{\footnotesize}
  \setlength{\abovecaptionskip}{4pt}
  \setlength{\belowcaptionskip}{-5pt}
  \centering
  \footnotesize
  \includegraphics[width=3.25in]{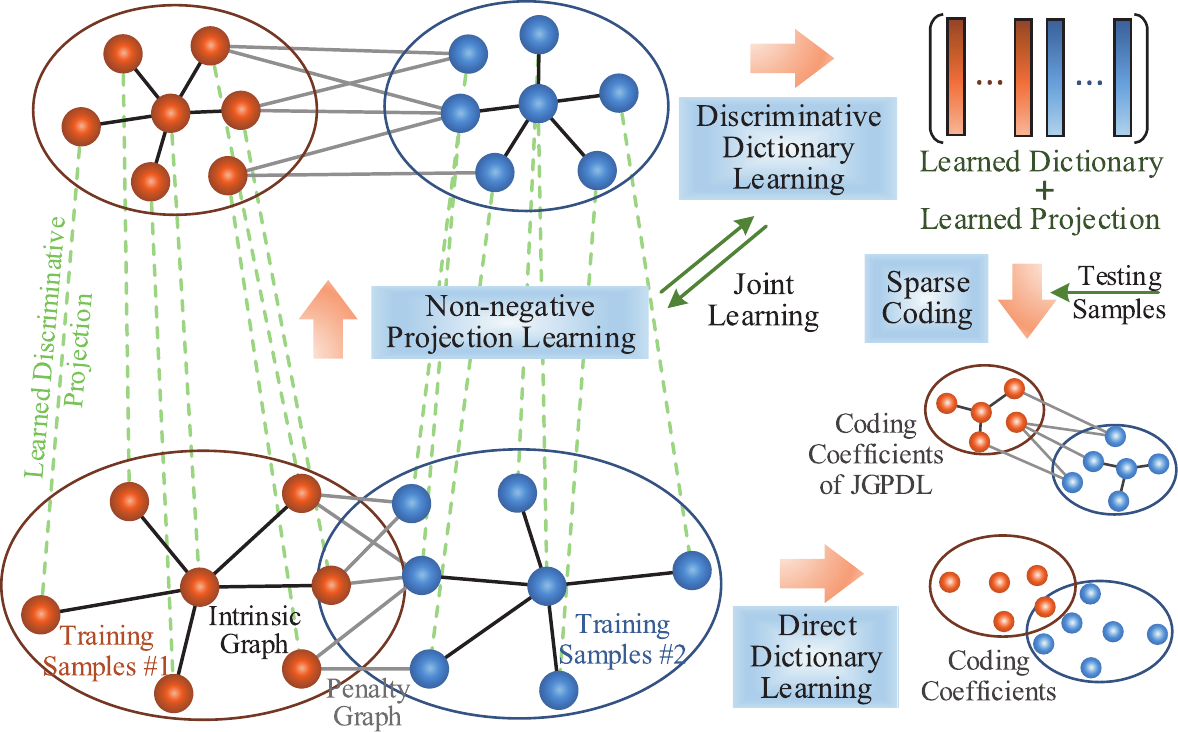}
  \caption{\footnotesize. An illustration of JNPDL. The dictionary and non-negative projection are jointly learned with discriminative graph constraints.}\label{fig1}
\end{figure}
\par
DL methods can be broadly divided into unsupervised DL and supervised DL. Due to the lack of label information, unsupervised DL only guarantees to discover signal patterns that are representative, but not necessarily discriminative. Supervised DL exploits the information among classes and requires the learned dictionary to be discriminative. Related literatures include the discriminative K-SVD \cite{zhang2010discriminative} and label-consistent K-SVD \cite{jiang2013label}. Dictionary regularization terms which takes label information into account were also introduced \cite{ramirez2010classification,castrodad2012sparse,qiu2014information}. For stronger discrimination, \cite{yang2011fisher,yang2014sparse} tried to learn a discriminative dictionary via the Fisher discrimination criteria. More recently, \cite{yang2014latent} proposed a latent DL method by jointly learning a latent matrix to adaptively build the relationship between dictionary atoms and class labels.
\par
Different from the conventional wide variety of discriminative DL literatures, our work casts an alternative view on this problem. One major purpose of this paper is to jointly learn a feature projection that improves DL. Instead of keep exploiting additional discrimination from the dictionary representation, we consider optimizing the input feature to further improve the learned dictionary. We believe such process can considerably influence the quality of learned dictionary, while a better learned dictionary may directly improve subsequent classification performance.
\par
Given that mid-level object parts are often discriminative for classification, we aim to learn a feature projection that mines these discriminative patterns. It is well-known that non-negative matrix factorization (NMF) \cite{lee1999learning} can learn similar part-like components. In the light of NMF and projective NMF (PNMF) \cite{liu2010projective}, we consider the projective self-representation (PSR) model where the set of training samples $\bm{Y}$ is approximately factorized as: $\bm{Y} \approx \bm{M}\bm{P}\bm{Y}$. The model jointly learns both the intermediate basis matrix $\bm{M}$ and the projection matrix $\bm{P}$ with non-negativity such that the additive (non-subtractive) combinations leads to learned projected features $\bm{P}\bm{Y}$ accentuating spatial object parts. In the paper, we propose a novel NMF-like feature projection learning framework on top of the PSR model to simultaneously incorporate label information with discriminative graph constraints. One shall see, our proposed framework can be viewed as a tradeoff between NMF and feature learning \cite{zou2012deep}.
\par
The dictionary representation is further discriminatively learned given the projected input features. An overview of the joint non-negative projection and dictionary learning (JNPDL) framework is illustrated in Fig. \ref{fig1}. The construction of discriminative graph constraints in both non-negative projection and dictionary learning follows the graph embedding framework \cite{yan2007graph}. While the inputs of graph constraints are essentially the same, they form different regularization terms for the convenience of optimization. Finally, a discriminative reconstruction constraint is also adopted so that coding coefficients will only well represent samples from their own classes but poorly represent samples from other classes. We test JNPDL in both image classification and image set classification with comprehensive evaluations, showing the excellent performance of JNPDL.

\section{Related Works}
Only a few works \cite{lu2014simultaneous,feng2013joint,zhang2013simultaneous} have discussed similar ideas but have all reported more competitive performance than conventional DL methods. \cite{lu2014simultaneous} proposed a framework which simultaneously learns feature and dictionary. Their work focused on learning a reconstructive feature (filterbank) with similar idea in \cite{zou2012deep}. The work in \cite{feng2013joint} jointly learns a dimensionality reduction matrix and a dictionary for face recognition. Unlike JNPDL, the model focused on low-dimensional representation without discriminative constraints. The source of discrimination purely comes from a Fisher-like constraint on coding coefficients. \cite{zhang2013simultaneous} presented a simultaneous projection and dictionary learning method with carefully designed discriminative sigmoid reconstruction error. Their method represents the input samples with the multiplication of projection matrix and dictionary, which differs significantly from JNPDL. \cite{gu2014projective} presents a novel dictionary pair learning approach for pattern classification, which also partially resembles our framework.

\section{The Proposed JNPDL Model}
Let $\bm{Y}$ be a set of $s$-dimensional training samples, i.e., $\bm{Y}=\{\bm{Y}_1,\bm{Y}_2,\cdots,\bm{Y}_K\}$ where $\bm{Y}_i$ denotes the training samples from class $i$. The learned structured (class-specific) dictionary is denoted by $\bm{D}=\{\bm{D}_1,\bm{D}_2,\cdots,\bm{D}_K\}$ and the corresponding sparse representation of the training samples over dictionary $\bm{D}$ is defined as $\bm{X}=\{\bm{X}_1,\bm{X}_2,\cdots,\bm{X}_K\}$. $\bm{M}$ is the intermediate non-negative basis matrix while $\bm{P}\in\mathbb{R}^{s_p\times s}$ denotes the projection matrix. In order to avoid high inter-class correlations and benefit subsequent sparse coding, JNPDL first projects training samples to a more discriminative space before dictionary encoding. Similar process applies for testing. At training, the projection, dictionary and encoded coefficients are jointly learned with the following model:
\begin{small}
\begin{equation}\label{eq1}
\begin{aligned}
\langle\bm{D},\bm{M},\bm{P}, \bm{X}\rangle &= \arg \min_{\bm{D},\bm{M} \geq\bm{0},\bm{P}\geq\bm{0},\bm{X}} \big{\{} R(\bm{D},\bm{P},\bm{X})+\alpha_1G_p(\bm{P},\bm{M})+\alpha_2G_c(\bm{X}) + \alpha_3\| \bm{X} \|_1\big{\}}
\end{aligned},
\end{equation}
\end{small}
where $\alpha_1$,$\alpha_2$ and $\alpha_3$ are scalar constants, and $R(\bm{D},\bm{P},\bm{X})$ is the discriminative reconstruction error. $G_p(\bm{P},\bm{M})$ is the graph-based projection term learning the NMF-like feature projection, while $G_c(\bm{X})$ is the graph-based coding coefficients term imposing discriminative label information to DL.
\subsection{Discriminative Reconstruction Error}
Discriminative reconstruction error targets the three following objectives: minimizing the global reconstruction error, minimizing the local reconstruction error\footnote{Stands for the reconstruction with atoms from the same class.} and maximizing the non-local reconstruction error. The term is defined as:
\begin{small}
\begin{equation}\label{eq2}
\begin{aligned}
R(\bm{D},\bm{P},\bm{X}) =\|\bm{P}\bm{Y}-\bm{D}\bm{X}\|_F^2+\sum_{i=1}^K\|\bm{P}\bm{Y}_i-\bm{D}_i\bm{X}^i_i\|_F^2+\sum_{i=1}^{K}\sum_{j=1,j\neq i}^{K}\|\bm{D}_j\bm{X}_i^j\|_F^2
\end{aligned},
\end{equation}
\end{small}
where $\bm{X}_i^j$ denotes the coding coefficients of samples $\bm{Y}_i$ associated with the sub-dictionary $\bm{D}_j$. $\sum_{i=1}^K\|\bm{Y}_i-\bm{D}_i\bm{X}^i_i\|_F^2$ denotes the local reconstruction error that requires the local sub-dictionaries $\bm{D}_i^i$ well represent samples from class $i$. $\|\bm{D}_j\bm{X}_i^j\|_F^2$ is further minimized such that inter-class coefficients $\bm{X}_i^j,i\neq j$ are relatively small compared with $\bm{X}_i^i$.
\subsection{Graph-based Coding Coefficients Term}
The term seeks to constrain the intra-class coding coefficients to be similar while the inter-class ones to be significantly dissimilar. We first construct an intrinsic graph for intra-class compactness and a penalty graph for inter-class separability. Rewriting the coding coefficients $\bm{X}$ as $\bm{X}=\{\bm{x}_1,\bm{x}_2,\cdots,\bm{x}_N\}$ where $N$ is the number of training samples, the similarity matrix of the intrinsic graph is defined as:
\begin{small}
\begin{equation}\label{eq3}
 \{\bm{W}_c\}_{ij} = \left\{ {\begin{array}{*{20}{l}}
{1,\ \ \text{\textnormal{if}}\ i\in S^+_{k_1}(j)\ \text{\textnormal{or}}\ j\in S^+_{k_1}(i)}\\
{0,\ \ \text{\textnormal{otherwise}}}
\end{array}} \right.,
\end{equation}
\end{small}
where $S^+_{k_1}(i)$ indicates the index set of the $k_1$ nearest intra-class neighbors of sample $\bm{x}_i$. Similarly, the similarity matrix of the penalty graph is defined as:
\begin{small}
\begin{equation}\label{eq4}
 \{\bm{W}_c^p\}_{ij} = \left\{ {\begin{array}{*{20}{l}}
{1,\ \ j\in S^-_{k_2}(i)\ \text{\textnormal{or}}\ i\in S^-_{k_2}(j)}\\
{0,\ \ \text{\textnormal{otherwise}}}
\end{array}} \right.,
\end{equation}
\end{small}
where $S^-_{k_2}(i)$ denotes the index set of the $k_2$ nearest inter-class neighbors of $\bm{x}_i$. the coefficients term is defined as:
\begin{small}
\begin{equation}\label{eq5}
\begin{aligned}
G_c(\bm{X})=\text{\textnormal{Tr}}(\bm{X}^T\bm{L}_c\bm{X})-\text{\textnormal{Tr}}(\bm{X}^T\bm{L}_c^p\bm{X})
\end{aligned},
\end{equation}
\end{small}
where $\bm{L}_c=\bm{B}_c-\bm{W}_c$ in which $\bm{B}_c=\sum_{j\neq i}\{\bm{W}_c\}_{ij}$ and $\bm{L}_c^p=\bm{B}_c^p-\bm{W}_c^p$ in which $\bm{B}_c^p=\sum_{j\neq i}\{\bm{W}_c^p\}_{ij}$. Imposing the graph-based discrimination makes the coding coefficients more discriminative. Interestingly, most existing discriminative coding coefficients term, such as in \cite{yang2014sparse,lu2014simultaneous}, are special cases of the graph-based discrimination constraint.

\subsection{Graph-based Non-negative Projection Term}
This term aims to learn a non-negative projection that maps the training samples to a discriminative space. Inspired by \cite{liu2010projective}, we design a structured projection matrix by dividing the projection matrix $\bm{P}$ into two parts:
\begin{small}
\begin{equation}\label{eq6}
\bm{Y}^{proj}=\begin{bmatrix}
\hat{\bm{Y}}^{proj} \\
\tilde{\bm{Y}}^{proj}
\end{bmatrix}=\bm{P}\bm{Y}=\begin{bmatrix}
\hat{\bm{P}} \\
\tilde{\bm{P}}
\end{bmatrix}\bm{Y}
\end{equation}
\end{small}
where $\hat{\bm{Y}}^{proj}=\{\hat{\bm{y}}^{proj}_1,\cdots,\hat{\bm{y}}^{proj}_N\}\in\mathbb{R}^{q\times N}$ that serves for the certain purpose of graph embedding, and $\tilde{\bm{Y}}^{proj}=\{\tilde{\bm{y}}^{proj}_1,\cdots,\tilde{\bm{y}}^{proj}_N\}\in\mathbb{R}^{(s_p-q)\times N}$ that contains the additional information for data reconstruction ($\tilde{\bm{Y}}^{proj}$ is a relaxed matrix that compensates the information loss in $\hat{\bm{Y}}^{proj}$). Note that $\hat{\bm{Y}}^{proj}$ preserves the discriminative graph properties while the whole $\bm{Y}^{proj}$ is used for data reconstruction purpose. Therefore the purposes of data reconstruction and graph embedding coexist harmoniously and do not mutually compromise like conventional formulations with multiple objectives. The basis matrix $\bm{M}$ is correspondingly divided into two parts $\bm{M}=\{\hat{\bm{M}},\tilde{\bm{M}}\}$ in which $\hat{\bm{M}}\in\mathbb{R}^{s\times q}$ and $\tilde{\bm{M}}\in\mathbb{R}^{s\times (s_p-q)}$. $\tilde{\bm{M}}$ can be considered as the complementary space of $\hat{\bm{M}}$.
\par
We first define $\bm{y}^{proj}_j$ as the $j$th column vector of $\bm{Y}^{proj}$ and then construct the intrinsic graph and the penalty graph using the same procedure as graph-based coding coefficients term. The construction of the similarity matrix $\bm{W}_p$ and $\bm{W}_p^p$ for intrinsic graph and the penalty graph is identical to $\bm{W}_c$ and $\bm{W}_c^p$, except that $\bm{W}_p$ and $\bm{W}_p^p$ measure the similarities among features and adopt different parameters. Differently, $\bm{W}_c$ and $\bm{W}_c^p$ measure similarities among coding coefficients. 
As \cite{yan2007graph} suggests, we have two objectives to preserve graph properties and enhance discrimination:
\begin{small}
\begin{equation}\label{eq7}
\left\{ {\begin{array}{*{20}{l}}
{\max_{\hat{\bm{Y}}^{proj}}\sum_{i\neq j} \|\hat{\bm{y}}^{proj}_i-\hat{\bm{y}}^{proj}_j\|^2_2\{\bm{W}_p^p\}_{ij}}\\
{\min_{\hat{\bm{Y}}^{proj}}\sum_{i\neq j} \|\hat{\bm{y}}^{proj}_i-\hat{\bm{y}}^{proj}_j\|^2_2\{\bm{W}_p\}_{ij}}
\end{array}} \right..
\end{equation}
\end{small}
\par
$\small \sum_{i\neq j}\|\bm{y}^{proj}_i-\bm{y}^{proj}_j\|^2_2\{\bm{W}_{p}\}_{ij}=\sum_{i\neq j}\|\hat{\bm{y}}^{proj}_i-\hat{\bm{y}}^{proj}_j\|^2_2\{\bm{W}_p\}_{ij}+\sum_{i\neq j}\|\tilde{\bm{y}}^{proj}_i-\tilde{\bm{y}}^{proj}_j\|^2_2\{\bm{W}_p\}_{ij}$, for a specific $\bm{Y}^{proj}$, minimizing the objective function with respect to $\tilde{\bm{Y}}^{proj}$ is equivalent to maximizing the objective function associated with the complementary part, i.e., $\hat{\bm{Y}}^{proj}$. Thus we constrain the projection matrix with the following equivalent objective:
\begin{small}
\begin{equation}\label{eq8}
\left\{ {\begin{array}{*{20}{l}}
{\min_{\hat{\bm{P}}}\text{\textnormal{Tr}}(\hat{\bm{P}}\bm{Y}\bm{L}_p\bm{Y}^T\hat{\bm{P}}^T)}\\
{\min_{\tilde{\bm{P}}}\text{\textnormal{Tr}}(\tilde{\bm{P}}\bm{Y}\bm{L}^p_p\bm{Y}^T\tilde{\bm{P}}^T)}
\end{array}} \right.
\end{equation}
\end{small}
where $\bm{L}_p=\bm{B}_p-\bm{W}_p$ in which $\bm{B}_p=\sum_{j\neq i}\{\bm{W}_p\}_{ij}$ and $\bm{L}_p^p=\bm{B}_p^p-\bm{W}_p^p$ in which $\bm{B}_p^p=\sum_{j\neq i}\{\bm{W}_p^p\}_{ij}$.  we formulate the graph-based projection term as follows:
\begin{small}
\begin{equation}\label{eq9}
\begin{aligned}
G_p(\bm{P},\bm{M})&=\|\bm{Y}-\bm{M}\bm{P}\bm{Y}\|_F^2+\beta\cdot\text{\textnormal{Tr}}(\hat{\bm{P}}\bm{Y}\bm{L}_p\bm{Y}^T\hat{\bm{P}}^T)+\beta\cdot\text{\textnormal{Tr}}(\tilde{\bm{P}}\bm{Y}\bm{L}^p_p\bm{Y}^T\tilde{\bm{P}}^T)+\|\bm{M}-\bm{P}^T\|_F^2
\end{aligned},
\end{equation}
\end{small}
where $\beta$ is a scaling constant and each column of $\bm{M}$ is normalized to unit $l_2$ norm. We use $\|\bm{Y}-\bm{M}\bm{P}\bm{Y}\|_F^2+\|\bm{M}-\bm{P}^T\|^2_F$ to incorporate the projection matrix into a NMF-like framework, and these two terms can further ensure the reconstruction ability of the projection $\bm{P}$ and avoid the trivial solutions of $\bm{P}$. they serve similar role to the auto-encoder style reconstruction penalty term in \cite{zuo2014learning,wang2014multi,zou2012deep}. Other terms in $G_p(\bm{P})$ preserve the graph properties and enhance discrimination.
\section{Optimization}
We adopt standard iterative learning to jointly learn $\bm{X}$, $\bm{P}$, $\bm{M}$ and $\bm{D}$. The proposed algorithm is shown in Algorithm \ref{alg1}. Because of the limited space, we present the detailed optimization framework in \textbf{Appendix A}. We also provide its convergence analysis in \textbf{Appendix B}. All the appendixes are included in the supplementary material which can be found at  \url{http://wyliu.com/}.
\par
{
\begin{algorithm}[t]
\small
    \caption{Training Procedure of JNPDL}
    \label{alg1}
    \begin{algorithmic}[1]
    \REQUIRE~Training samples $\thickmuskip=5mu \bm{Y}=\|\bm{Y}_1,\cdots,\bm{Y}_N\|$, intrinsic graph $\bm{W}_c,\bm{W}_p$, penalty graph $\bm{W}_c^p,\bm{W}_p^p$, parameters $\alpha_1,\alpha_2,\alpha_3,\beta$, iteration number $T$.
    \ENSURE~Non-negative projection matrix $\bm{P}$, dictionary $\bm{D}$, coding coefficient matrix $\bm{X}$.\\
    \textbf{Step1: Initialization}
    \STATE $t=1$.
    \STATE Randomly initialize columns in $\bm{D}^0,\bm{M}^0$ with unit $l_2$ norm.
    \STATE Initialize $\bm{x}_{i,1\leq i\leq N}$ with $((\bm{D}^0)^T(\bm{D}^0)+\lambda_2\bm{I})^{-1}(\bm{D}^0)^T\bm{y}_i$\ where $\bm{y}_i$ is the $i$th training sample (regardless of label).\\
    \textbf{Step2: Search local optima}
    \WHILE {not convergence or $t<T$}
    \STATE Solve $\bm{P}^t,\bm{M}^t$ iteratively with fixed $\bm{D}^{t-1}$ and $\bm{X}^{t-1}$ via non-negative projection learning.\\
    \STATE Solve $\bm{X}^t$ with fixed $\bm{M}^t$, $\bm{D}^{t-1}$ and $\bm{P}^t$ via non-negative projection learning.\\
    \STATE Solve $\bm{D}^t$ with fixed $\bm{M}^t$, $\bm{P}^{t}$ and $\bm{X}^t$ via discriminative DL.\\
    \STATE $t\leftarrow t+1$.
    \ENDWHILE
    \\\textbf{Step3: Output}
    \STATE Output $\bm{P}=\bm{P}^t$, $\bm{D}=\bm{D}^t$ and $\bm{X}=\bm{X}^t$.
    \end{algorithmic}
 \end{algorithm}
}

\section{Classification Strategy}
When the projection and the dictionary have been learned, we need to project the testing image via learned projection, code the projected sample over the learned dictionary and eventually obtain its coding coefficients which is expected to be discriminative. We first project the testing sample and then code it over the learning dictionary via
\begin{small}
\begin{equation}\label{cla}
\hat{\bm{x}}=\arg\min_{\bm{x}}\{\|\bm{P}\bm{y}-\bm{D}\bm{x}\|_2^2+\lambda_1\|\bm{x}\|_1\}
\end{equation}
\end{small}
where $\lambda$ is a constant. After obtaining the coding coefficients $\hat{\bm{x}}$, we classify the testing sample via
\begin{small}
\begin{equation}\label{cla2}
label(\bm{y})=\arg\min_{i}\big{\{}\|\bm{P}\bm{y}-\bm{D}_i\delta_i(\bm{x})\|_2^2+\sigma\|\bm{x}-\bm{m}_i\|_2^2\big{\}}
\end{equation}
\end{small}
where $\sigma$ is a weight to balance these two terms, and $\delta_i(\cdot)$ is the characteristic function that selects the coefficients associated with $i$th class. $\bm{m}_i$ is the mean vector of the learned coding coefficient matrix of class $i$, i.e., $\bm{X}_i$. Incorporating the term $\|\bm{x}-\bm{m}_i\|_2^2$ is to make the best of the discrimination within the dictionary, because the dictionary is learned to make coding coefficients similar from the same class and dissimilar among different classes.
\section{Experiments}
\subsection{Implementation Details}
We evaluate JNPDL by both image classification and image set classification. We construct $\bm{W}_p$ and $\bm{W}_p^p$ with correlation similarity, and set the number of nearest neighbors as $\min(n_c-1,5)$ where $n_c$ is the training sample number in class $c$. The number of shortest pairs from different classes is 20 for each class. For $\bm{W}_c$ and $\bm{W}_c^p$, we first remove the graph-based coding coefficients term in Eq. \ref{eq1} and solve the optimization. The Euclidean distances among the coding coefficients of training samples are used as initial neighbor metric. We set $k_1=\min(n_c-1,5),k_2=30$. For all experiments, we fix $\alpha_1=1,\alpha_2=1$ and $\alpha_3=0.05$. Other parameters for JNPDL are obtained via 5-fold cross validation to avoid over-fitting. Specifically, we use $\beta=0.7, \lambda_1=5\times10^{-6}$ and $\sigma=0.05$ for image classification. For image set based face recognition, we set $\beta=0.8, \lambda_2=0.001\times N/700$ where $N$ is the number of training samples. For all baselines, we usually use their original settings or carefully implement them following the paper. All results are the average value of 10 times independent experiments.
\subsection{Application to Image Classification}
\subsubsection{Face recognition}
We evaluate JNPDL on the Extended Yale B (ExYaleB)\footnote{The extended Yale B dataset consists of 2,414 frontal face images from 38 individuals. All images are normalized to $54\times 48$.} and AR Face Dataset\footnote{The AR dataset consists of over 4,000 frontal images for 126 individuals. For each individual, 26 pictures were taken in two separated sessions. All images in AR are normalized to $60\times 43$.}. For both ExYaleB and AR Face, we exactly follow the same training/testing set selection in \cite{yang2011fisher}. The dictionary size is set to half of the training samples. SRC \cite{wright2009robust} uses all training samples as the dictionary.
\par
\textbf{Comparison with state-of-the-art approaches.} We compare JNPDL with state-of-the-art DL approaches including D-KSVD \cite{zhang2010discriminative}, LC-KSVD \cite{jiang2013label} and FDDL \cite{yang2014latent}. DSRC \cite{zhang2013simultaneous} and JDDRDL \cite{feng2013joint} which share similar philosophy are also compared. JNPDL, DSRC and JDDRDL uses the original images for training and set the feature dimension after projection as 300. All the other methods use the 300-dimensional Eigenface feature. SRC and Linear SVM are used as baselines. Results are shown in Table \ref{acctab}. One can see that JNPDL achieves promising recognition accuracy on both datasets, respectively achieving 2.2\% and 2.6\% improvement over the second best approaches.
\par
\begin{table}[t]
\renewcommand{\captionlabelfont}{\footnotesize}
\setlength{\abovecaptionskip}{2pt}
\setlength{\belowcaptionskip}{0pt}
\centering
\footnotesize
\caption{\footnotesize . Recognition accuracy (\%) on extended Yale B and AR.}\label{acctab}
\begin{tabular}{|c|c|c|c|c|c|}
\hline
\textbf{Method} & \textbf{ExYaleB} & \textbf{AR Face} & \textbf{Method} & \textbf{ExYaleB} & \textbf{AR Face} \\
\hline\hline
SVM & 88.8 & 87.1 & JDDRDL \cite{feng2013joint} & 90.1 & 90.9 \\
SRC \cite{wright2009robust} & 91.0 & 88.5 &FDDL \cite{yang2014sparse} & 91.9 & 92.1 \\
D-KSVD \cite{zhang2010discriminative} & 75.3 & 85.4 & DSRC \cite{zhang2013simultaneous} & 89.6 & 88.2 \\
LC-KSVD \cite{jiang2013label} & 86.8 & 90.2 & \textbf{JNPDL} & \textbf{94.1} & \textbf{94.7} \\
\hline
\end{tabular}
\end{table}
\par
\textbf{Accuracy vs. Feature dimensionality.} We vary the feature dimension after projection to evaluate the performance of JNPDL on AR dataset. For SRC and FDDL, the dimensionality reduction is performed by Eigenface. Table \ref{dimtab} indicates that jointly learned projection can preserve much discriminative information even at low feature dimension.
\par
\begin{table}[t]
\renewcommand{\captionlabelfont}{\footnotesize}
  \setlength{\abovecaptionskip}{4pt}
  \setlength{\belowcaptionskip}{-5pt}
\centering
\footnotesize
\caption{\footnotesize . Accuracy (\%) vs. feature dimension on AR dataset.}\label{dimtab}
\begin{tabular}{|c|c|c|c|c|}
\hline
\textbf{Dimension} & \textbf{100} & \textbf{200} & \textbf{300} & \textbf{500} \\
\hline\hline
SRC \cite{wright2009robust} & 84.0 & 87.3 & 88.5 & 89.7 \\
FDDL \cite{yang2014sparse} & 85.7 & 88.5 & 92.0 & 92.2 \\
DSRC \cite{zhang2013simultaneous} & 84.8 & 86.9 & 88.2 & 89.1 \\
JDDRDL \cite{feng2013joint} & 82.5 & 87.7 & 90.9 & 91.6 \\
\textbf{JNPDL} & \textbf{88.3} & \textbf{92.4} & \textbf{94.7} & \textbf{95.1} \\
\hline
\end{tabular}
\end{table}
\par

\textbf{Joint projection learning vs. Separate projection.} Projection and dictionary can also be learned separately. We compare it with joint learning to validate our motivation for JNPDL. We also remove projection learning from JNPDL and use Eigenface features with dimension set to 300. Results are shown in Table \ref{jstab}. Results show that JNPDL jointly learning projection and dictionary achieves the best accuracy.
\par
\begin{table}[t]
\renewcommand{\captionlabelfont}{\footnotesize}
  \setlength{\abovecaptionskip}{4pt}
  \setlength{\belowcaptionskip}{-5pt}
\centering
\footnotesize
\caption{\footnotesize . Accuracy (\%) of different projection.}\label{jstab}
\begin{tabular}{|c|c|c|}
\hline
\textbf{Method} & \textbf{ExYaleB} & \textbf{AR Face} \\
\hline\hline
JNPDL (with Eigenface) & 91.8 & 92.2 \\
JNPDL (Separate Learning) & 92.1 & 92.5 \\
JNPDL (Joint Learning) & \textbf{94.1} & \textbf{94.7} \\
\hline
\end{tabular}
\end{table}
\par

\textbf{Face recognition in the wild.} we apply JNPDL in a more challenging face recognition task with LWFa dataset \cite{wolf2011effective} which is an aligned version of LFW. We use 143 subject with no less than 11 samples per subject in LFWa dataset (4174 images in total) to perform the experiment. The first 10 samples are selected as the training samples and the rest is for testing. Following \cite{yang2014latent}, histogram of uniform-LBP is extracted by partitioning face into $10\times 8$ patches. PCA is used to reduce the dimension to 1000. Results are shown in Table \ref{lfwatab}, with JNPDL achieving the best.
\par
\begin{table}[t]
\renewcommand{\captionlabelfont}{\footnotesize}
  \setlength{\abovecaptionskip}{4pt}
  \setlength{\belowcaptionskip}{-5pt}
\centering
\footnotesize
\caption{\footnotesize . Recognition Accuracy (\%) on LFWa Dataset.}\label{lfwatab}
\begin{tabular}{|c|c|c|c|c|c|c|c|}
\hline
\textbf{Method} & \textbf{Acc.} & \textbf{Method} & \textbf{Acc.}& \textbf{Method} & \textbf{Acc.}& \textbf{Method} & \textbf{Acc.}\\
\hline\hline
SVM & 63.0 & COPAR \cite{kong2012dictionary} & 72.6 & D-KSVD & 65.9 & LDL \cite{yang2014latent} & 77.2 \\
SRC & 72.7 & FDDL & 74.8 & LC-KSVD & 66.0 & \textbf{JNPDL} & \textbf{78.1} \\
\hline
\end{tabular}
\end{table}
\par
\subsubsection{Object categorization}
We perform the object categorization experiment on 17 Oxford Flower dataset and use the default experiment setup as in \cite{yang2014sparse}. We compare JNPDL with MTJSRC \cite{yuan2012visual}, COPAR, JDDRDL, DSRC, FDDL, SDL, LLC and two baseline: SRC, SVM. For fair comparison, we use the Frequent Local Histogram (FLH) feature to generate a kernel-based feature descriptor the same as \cite{yang2014sparse}. Table \ref{floacc} shows that JNPDL is slightly worse than FDDL but is better than most competitive approaches. We believe that it is because the kernel features are already quite discriminative and projection does not help much.
\begin{table}[t]
\renewcommand{\captionlabelfont}{\footnotesize}
  \setlength{\abovecaptionskip}{4pt}
  \setlength{\belowcaptionskip}{-5pt}
\centering
\footnotesize
\caption{\footnotesize . Recognition Accuracy (\%) on 17 Oxford Flower Dataset.}\label{floacc}
\begin{tabular}{|c|c|c|c|}
\hline
\textbf{Method} & \textbf{Accuracy} & \textbf{Method} & \textbf{Accuracy}\\
\hline\hline
SVM & 88.6 & MTJSRC & 88.4 \\
SRC & 88.4 & COPAR & 88.6 \\
LLC (20 bases) & 89.7 & SDL & 91.0 \\
JDDRDL & 87.7 & FDDL & 91.7 \\
DSRC & 88.9 & \textbf{JNPDL} & \textbf{92.1} \\
\hline
\end{tabular}
\end{table}
\subsection{Application to Image Set Classification}
\subsubsection{Classification strategy for image set classification}
Applying the classification in Section 5 to each video frame altogether with a voting strategy, JNPDL can be easily extended to image set classification. Given a testing video $Y^{te}=\{\bm{y}_1^{te},\bm{y}_2^{te},\cdots,\bm{y}_{K_t}^{te}\}$ in which $\bm{y}_j^{te}$ is the $j$th frame and $K_t$ is the number of image frames in the video, we project each frame to a feature via the learned non-negative projection $\bm{P}$ and obtain its coding coefficients with Eq. \eqref{cla}. Thus the label of a video frame can be obtain by Eq. \eqref{cla2}. After getting all the labels of frames, we perform a majority voting to decide the label of the given image set. For testing efficiency, we replace the $l_1$ norm $\|\bm{x}\|_1$ with a $l_2$ norm $\|\bm{x}\|_2^2$ and derive the decision:
\begin{small}
\begin{equation}\label{cla3}
label(\bm{y}_j^{te})=\arg\min_{i}\{\|\bm{P}\bm{y}_j^{te}-\bm{D}_i\delta_i(\bm{D}^{\dagger}\bm{y}_j^{te})\|_2^2\}.
\end{equation}
\end{small}
where $\bm{D}^{\dagger}=(\bm{D}^T\bm{D}+\lambda_2\bm{I})^{-1}\bm{D}^T$. Eventually we use the majority voting to decide the label of a video (image set).
\subsubsection{Image set based face recognition}
Three video face recognition benchmark dataset, including Honda/UCSD \cite{lee2003video}\footnote{Contains 59 face video of 20 individuals with large pose and expression variations, and average length of 400 frames.}, CMU MoBo \cite{gross2001cmu}\footnote{Consists of 96 videos from 24 subjects, each containing 5 videos of different walking patterns.} and YouTube Celebrities (YTC)\footnote{Contains 1910 video sequences of 47 celebrities from YouTube. Most videos contain noisy and low-resolution image frames.} are used to evaluate the proposed JNPDL. For fair comparison, we follows the experimental setup in \cite{lu2014simultaneous}. We use Viola-Jones face detector to capture faces and then resize them to $30\times 30$ intensity image. Each image frame is cropped into $30\times 30$ according to the provided eye coordinates. Thus each video is represented as an image set. Following standard experiment protocol as in \cite{cevikalp2010face,lu2014simultaneous}, the detected face images are histogram equalized but no further preprocessing, and the image features are raw pixel values.
\par
\textbf{Comparison with state-of-the-art approaches.} For both the Honda/UCSD and CMU MoBo datasets, we randomly select one face video per person as the training samples and the rest as testing samples. For YTC dataset, we equally divide the whole dataset into five folds, and each fold contains 9 videos per person. In each fold, we randomly select 3 face videos per person for training and use the rest for testing. We compare JNPDL with DCC \cite{kim2007discriminative}, MMD \cite{wang2008manifold}, MDA \cite{wang2009manifold}, CHISD \cite{cevikalp2010face}, SANP \cite{hu2011sparse}, LMKML \cite{lu2013image} and SFDL \cite{lu2014simultaneous}. The settings of these approaches are basically the same as \cite{lu2014simultaneous}. We select the best accuracy that JNPDL achieves with projected dimensions from 50, 100, 150, 200 and 300. Results in Table \ref{acctab2} show the superiority of the proposed method.
\begin{table}[t]
\renewcommand{\captionlabelfont}{\footnotesize}
\setlength{\abovecaptionskip}{4pt}
\setlength{\belowcaptionskip}{0pt}
\centering
\footnotesize
\caption{\footnotesize . Recognition acc. (\%) on Honda, MoBo, YTC datasets.}\label{acctab2}
\begin{tabular}{|c|c|c|c|c|c|c|c|}
\hline
\textbf{Method} & \textbf{Honda} & \textbf{MoBo} & \textbf{YTC} & \textbf{Method} & \textbf{Honda} & \textbf{MoBo} & \textbf{YTC} \\
\hline\hline
DCC \cite{kim2007discriminative}& 94.9 & 88.1 & 64.8 & SANP \cite{hu2011sparse} & 93.6 & 96.1 & 68.3 \\
MMD \cite{wang2008manifold} & 94.9 & 91.7 & 66.7 & LMKML \cite{lu2013image} & 98.5 & 96.3 & \textbf{78.2} \\
MDA \cite{wang2009manifold} & 97.4 & 94.4 & 68.1 & SFDL \cite{lu2014simultaneous} & 100 & 96.7 & 76.7 \\
CHISD \cite{cevikalp2010face} & 92.5 & 95.8 & 67.4 & \textbf{JNPDL} & \textbf{100} & \textbf{97.1} & \textbf{77.4}\\
\hline
\end{tabular}
\end{table}
\begin{figure}[t!]
  \renewcommand{\captionlabelfont}{\footnotesize}
  \setlength{\abovecaptionskip}{4pt}
  \setlength{\belowcaptionskip}{0pt}
  \centering
  \includegraphics[width=2.9in]{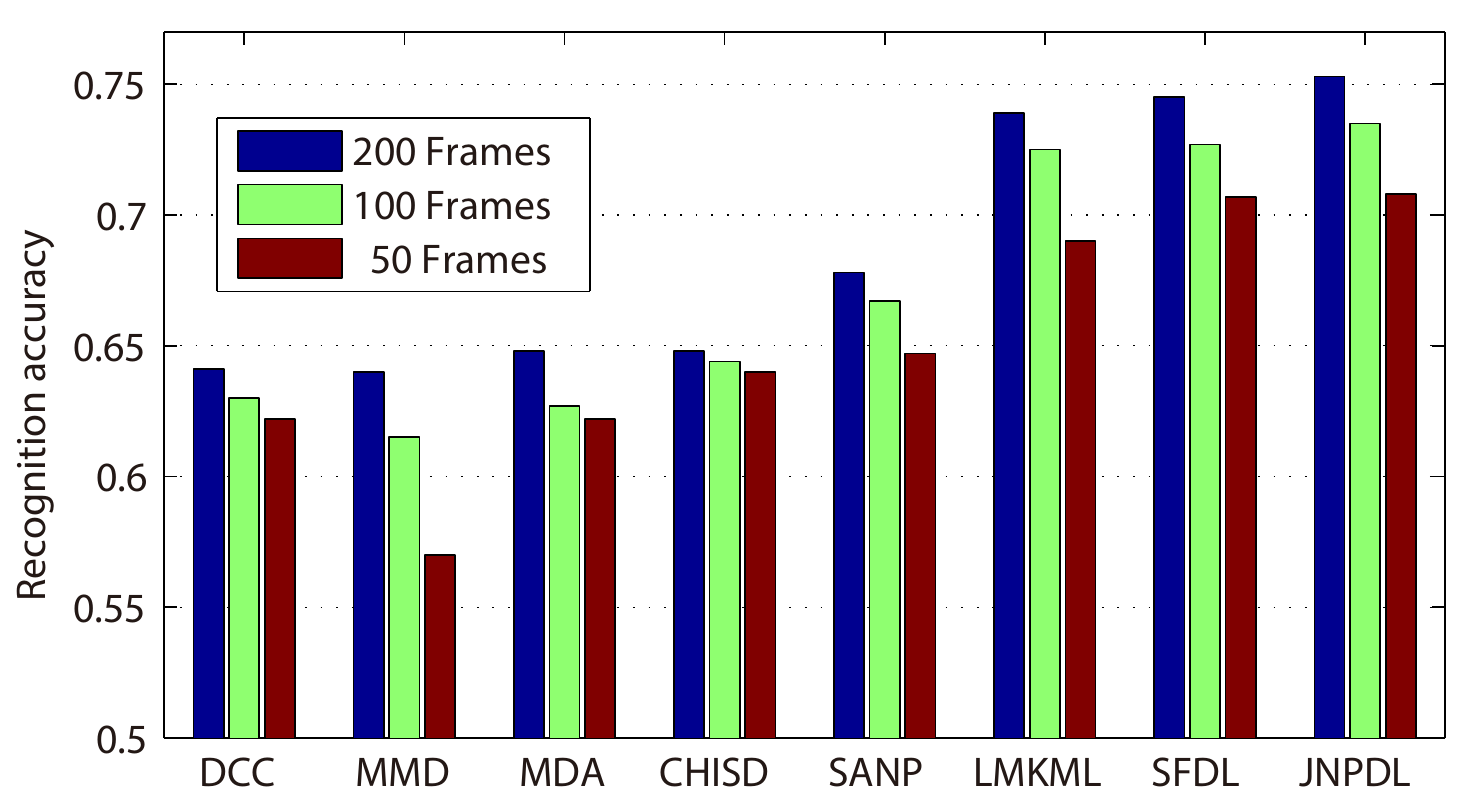}
  \caption{\footnotesize. Recognition accuracy with different number of frames.}\label{fig4}
\end{figure}
\par
\textbf{Accuracy vs. Frame number.} We evaluate the performance of JNPDL when videos contain different number of image frames on YTC dataset. We randomly select 50, 100 and 200 frames from each face image set for training and select another 50, 100, 200 for testing. If an image set does not have enough frames, all frames are used. We select the best accuracy JNPDL achieves with projected dimension equal to 50, 100, 150, 200 and 300. Fig. \ref{fig4} shows JNPDL achieves better accuracy than the other approaches. It can be learned that the discrimination power of JNPDL is strong even with small training set.
\par
\textbf{Efficiency.} The computational time for JNPDL to recognize a query image set is approximately 5 seconds with a 3.4GHz Dual-core CPU and 16GB RAM, which is comparable to \cite{lu2014simultaneous,cevikalp2010face} and slightly higher than \cite{kim2007discriminative,wang2008manifold,wang2009manifold}.
\section{Concluding Remarks}
In this paper, we proposed a novel joint non-negative projection and dictionary learning framework where non-negative feature projection and dictionary are simultaneously learned with discriminative graph constraints. The graph constraints guarantee the discrimination of projected training samples and coding coefficients. We also proposed a multiplicative non-negative updating algorithm for the projection learning with convergence guarantee. The learned feature projection considerably improves the quality learned dictionary, leading to better classification performance. Experimental results have validated the excellent performance of JNPDL on both image classification and image set classification.
\par
Possible future work includes handling nonlinear cases using methods like kernel trick or other non-linear mapping algorithms, adding more discriminative regularizations to learn the projection matrix and considering to learn a multiple-layered (hierarchical) projection.
\section{Acknowledgement}
This work is partially supported by the National Natural Science Foundation for Young Scientists of China (Grant  no.  61402289), National  Science  Foundation  of  Guangdong  Province  (Grant  no.  2014A030313558)
{
\newcommand{\qihao}{\fontsize{8pt}{\baselineskip}\selectfont}
\small
\bibliographystyle{named}
\bibliography{egbib}
}
\end{document}